\documentclass{article}

     \PassOptionsToPackage{numbers, compress}{natbib}

\usepackage{amsmath}
\usepackage{multirow}
\usepackage{booktabs}
\usepackage{amssymb}
\usepackage{graphicx}
\usepackage{enumitem}

\usepackage[preprint]{2025}

\usepackage[utf8]{inputenc} 
\usepackage[T1]{fontenc}    
\usepackage{hyperref}       
\usepackage{url}            
\usepackage{booktabs}       
\usepackage{amsfonts}       
\usepackage{nicefrac}       
\usepackage{microtype}      
\usepackage{xcolor}         

\title{Refining CNN-based Heatmap Regression with Gradient-based Corner Points for Electrode Localization}

\author{Lin Wu}

\begin{document}

\maketitle

\begin{abstract}
We propose a method for detecting the electrode positions in lithium-ion batteries. The process begins by identifying the region of interest (ROI) in the battery's X-ray image through corner point detection. A convolutional neural network is then used to regress the pole positions within this ROI. Finally, the regressed positions are optimized and corrected using corner point priors, significantly mitigating the loss of localization accuracy caused by operations such as feature map down-sampling and padding during network training. Our findings show that combining traditional pixel gradient analysis with CNN-based heatmap regression for keypoint extraction enhances both accuracy and efficiency, resulting in significant performance improvements.
\end{abstract}

\section{Introduction}
The quality of battery cells is directly linked to the precise localization and alignment of the electrodes.~\cite{du2022impact} The folding of the electrode sheets or poor alignment of the four corners of the electrode assembly in lithium-ion batteries can lead to reduced performance, overheating risks, mechanical damage, and safety hazards. Therefore, ensuring the accurate alignment and shape of the electrode sheets during the manufacturing process is crucial, particularly for lithium-ion batteries used in high-performance and high-safety applications, such as electric vehicles and energy storage systems.

However, due to the multi-levels structure of the cells and the significant overlap between the electrode sheets, X-ray images often suffer from low brightness, high noise, and variations in pole area sizes and positions~\cite{masuch2022applications}. These challenges make accurate electrode detection particularly difficult. Existing automated detection algorithms fail to fully leverage the unique features of X-ray images, leading to a high rate of overkill. To this end, we propose a novel joint optimization framework that incorporates a gradient-based corner point detection method to refine CNN-based heatmap regression for electrode localization.

Our contributions are summarized as follows:

\begin{itemize}
\item The region of interest (ROI) corresponding to the pole area is efficiently identified in high-resolution X-ray images using corner point detection, that is, Oriented Features from Accelerated Segment Test (OFAST)~\cite{rublee2011orb}.

\item A high-resolution neural network (HRNet)~\cite{sun2019deep} with heatmap regression is employed to detect pole coordinates, enhancing model adaptability through various data augmentation techniques.

\item A confidence-based adjustment module is introduced to dynamically refine pole detection results by evaluating the confidence of the poles predicted by HRNet and their reference corners detected by OFAST.

\item Various evaluation metrics validate the effectiveness of our joint optimization framework, which combines CNN-based heatmap regression~\cite{payer2016regressing} and gradient-based corner point detection. Additionally, PCS (Percentage of Correct Samples) is introduced as a novel metric to calculate the proportion of samples with maximum normalized error below a threshold.
\end{itemize}

\section{Related Work}
\paragraph{Electrode Localization in X-ray Images.}
Precise electrode localization is essential for quality assurance in lithium-ion battery manufacturing, particularly in safety-critical applications such as electric vehicles, where misaligned electrodes can pose significant hazards~\cite{masuch2022applications, etiemble2015quality}. Traditional methods, including Canny edge detection~\cite{rong2014improved} and Hough transforms~\cite{zhang2022comprehensive}, assume clear geometric boundaries and fail under the low contrast and heavy overlap characteristic of real-world industrial X-ray images. Recent learning-based approaches have focused on coarse-grained tasks such as defect classification\cite{zuo2023x} or component segmentation~\cite{saberironaghi2023defect}, but largely overlook the fine-grained localization required for accurate pole coordinate estimation. Our work bridges this gap by introducing a novel approach for high-precision pole coordinate detection.

\paragraph{Heatmap Regression with Geometric Priors.}
Heatmap regression is widely adopted for keypoint detection across domains, from human pose estimation~\cite{wang2021deep} to medical image analysis~\cite{wu2023key}, with HRNet~\cite{sun2019deep} achieving state-of-the-art performance via consistent high-resolution representations. However, two challenges arise in the context of X-ray electrode localization: (1) weak boundary gradients result in diffuse or ambiguous heatmaps, as also noted in facial landmark detection~\cite{watchareeruetai2022lotr}; and (2) conventional CNNs lack awareness of global geometric structure. To overcome these limitations, we propose a hybrid framework that combines learnable heatmap prediction with deterministic corner proposals generated by gradient-based detectors, thereby incorporating explicit geometric priors into the localization process.

\paragraph{Gradient-Driven Corner Detection.}
Classical corner detectors such as Harris, FAST~\cite{biadgie2014feature}, and OFAST~\cite{rublee2011orb} remain valuable in industrial imaging due to their robustness under noise—an advantage that has been underexploited in recent deep learning pipelines. Unlike segmentation-driven methods~\cite{saberironaghi2023defect}, we employ OFAST not as a standalone detector, but as a geometric regularizer that guides and refines CNN-based localization. This strategy parallels recent work on hybrid vision systems for document layout analysis~\cite{semma2021writer}, but is adapted here to address electrode-specific challenges such as multi-layer interference and occlusion.

\section{Methodology}
\begin{figure}[htb]
\centerline{\includegraphics[width=\linewidth]{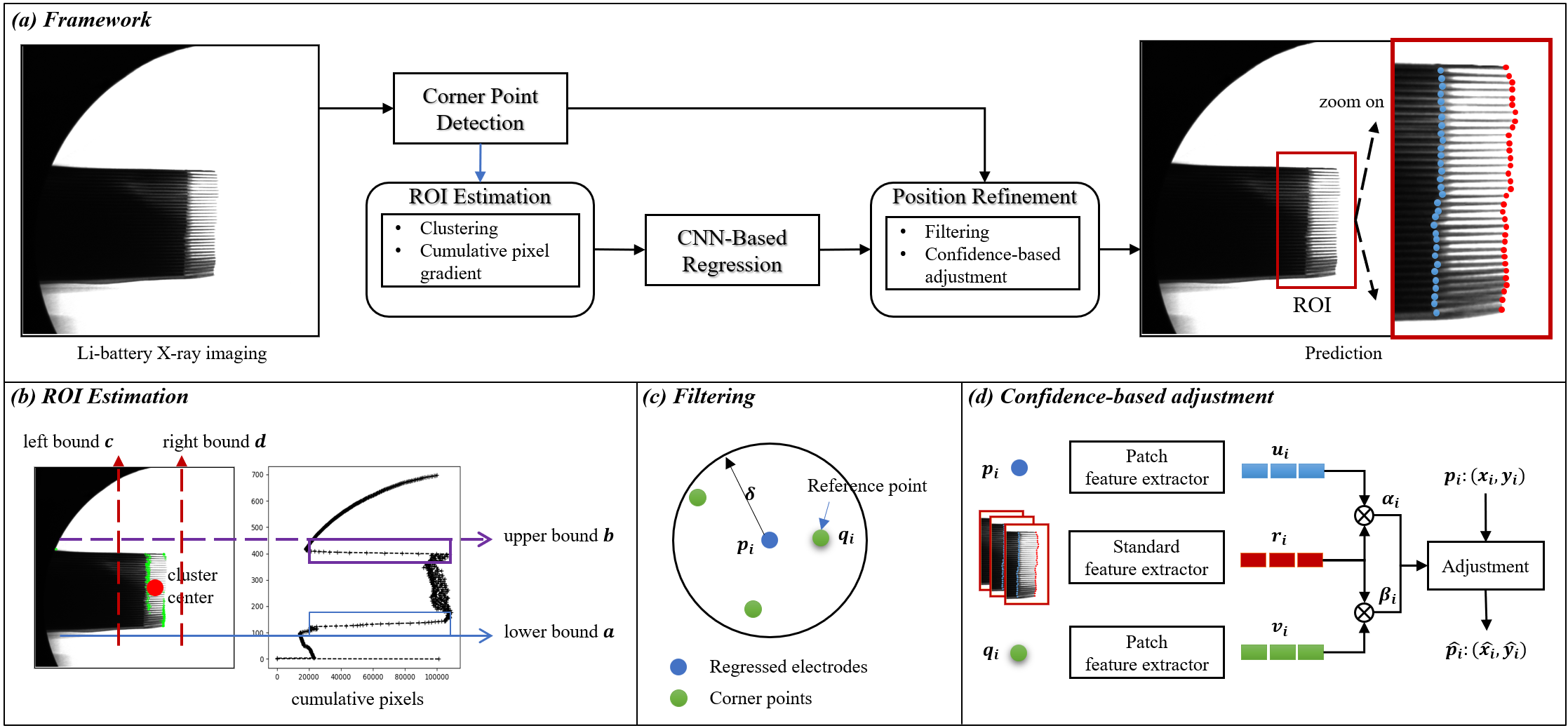}}
\caption{Framework of the proposed Electrode Localization and Optimization}
\label{fig:framework}
\end{figure}

Our framework for electrode localization and optimization is illustrated in the Fig.~\ref{fig:framework} (a). Given high-resolution X-ray images of the lithium-ion battery, we initially employ a corner point detection model to extract a set of corner points. These corner points are subsequently input into an ROI estimation module to identify a smaller area for further focus. Within this ROI, pole positions are detected using a CNN-based heatmap regression model. Finally, the pole positions predicted by the CNN model are refined and corrected by leveraging the corner points previously identified through gradient-based corner detection.

\subsection{Corner Points and ROI Detection}

OFAST is an improvement on the traditional FAST (Features from Accelerated Segment Test) algorithm, aiming to enhance its rotation invariance. The core idea of OFAST is to detect corner points in an image based on local brightness changes and calculate the direction of each corner point, so that the algorithm can still maintain good matching results when facing rotation or posture changes. Due to the fact that X-ray images of lithium batteries are primarily grayscale images, and the brightness gradient features at the pole positions are very rich, we therefore use OFAST to extract corner points.

As shown in Fig.~\ref{fig:framework} (b), these detected corner points (in green) are primarily concentrated in the pole region of the lithium battery, although they do not perfectly match the actual pole positions. It can be observed that at least the actual pole positions must have distinct brightness gradients. This phenomenon encourages us to make full use of the distribution information of these corner points. Specifically, let the detected set of corner points be denoted as $Q={\{q_1,q_2,...,q_n\}}, q_i \in \mathbb{R}^2$, we first calculate their cluster center (only one cluster) as follows:
\begin{equation}
    x_0,\  y_0 = \frac{1}{n} \sum_{i=1}^{n} q_{i}
\end{equation}
After obtaining the center point, we accumulate the row-wise grayscale values of the X-ray image to form a new "pattern," as shown in Fig.~\ref{fig:framework} (b). This pattern enables us to easily identify the upper and lower bounds of the ROI, which correspond to the locations where the row-wise accumulated gradient experiences a significant change. We set a threshold \( \tau \) to detect these gradient changes, and define two pointers: the upper pointer moves from 0 to \( y_0 \) with a step size \( \Delta b \), while the lower pointer moves from the image height \( H \) to \( y_0 \) with a step size \( \Delta a \). This process can be formalized as follows:
\begin{equation}
    b = 0 + n \Delta b, \quad n = 0, 1, 2, \dots, \quad a = H - m \Delta a, \quad m = 0, 1, 2, \dots
\end{equation}

where \( n \) and \( m \) are the step indices, and \( \Delta b \) and \( \Delta a \) are the step sizes for the upper and lower pointers, respectively.
Once we have the upper bound \( b \) and the lower bound \( a \), we introduce a new variable \( \lambda \) to control the width of the ROI, which is determined by the horizontal coordinate of the cluster center and the height of the ROI. We therefore achieve the left bound $c$ and right bound $d$ as follows:
\begin{align}
    c = x_0 - \lambda (b - a)  \\
    d = x_0 + \lambda (b - a)
\end{align}
where \( \lambda \) is a scaling factor that adjusts the width of the ROI based on the cluster center’s horizontal position and the height of the ROI.

\subsection{Electrode Regression Network}
Once we have the ROI, we can use any keypoint detection method to estimate the poles in the image. Due to the significant advantages of HRNet in high-resolution feature extraction and precise keypoint localization, it has been widely adopted for high-accuracy keypoint detection tasks. Additionally, HRNet utilizes heatmap regression to locate keypoints, which enhances the model's accuracy in keypoint detection. Therefore, we choose HRNet as our electrode regression network. 

During training, we applied data augmentation techniques such as rotation, reflection, and other affine transformations. The heatmap regression network is trained using Mean Squared Error (MSE) as the loss function. The training set consists of approximately 2,000 samples, while the test set contains around 1,000 samples.

\subsection{Post-Correction with Corner Points} 
Due to the computational efficiency and hardware deployment considerations of heatmap regression-based 2D keypoint estimation models, the resolution of the feature maps is typically lower than the original input resolution. When determining the final keypoint coordinates, the feature map resolution is upsampled to match the original resolution. For example, in HRNet, the feature map scale is 1/4 of the input resolution. Additionally, operations such as multi-layer downsampling, upsampling, and padding inevitably introduce some discrepancy between the predicted and ground truth keypoint coordinates. To further reduce this discrepancy, we propose a post-correction algorithm based on corner point detection.

Specifically, for each predicted pole coordinate \( p_i \) from the regression network, the radius \( \delta \) is set, and the nearest corner point within the circle of radius \( \delta \) around \( p_i \) is selected as the reference corner point \( q_i \), as shown in Fig.~\ref{fig:framework} (c). Note that if no corner points exist within the radius \( \delta \), the point \( p_i \) will not have a reference corner. Although we have captured the \( p_i, q_i \) pairs, inevitable noise or errors may still affect them, which is caused by neural network training and local brightness gradient patterns. Therefore, we need to evaluate the confidence of the two pole position estimates before a potential fusion and interaction, as shown in Fig.~\ref{fig:framework} (d).

Before evaluating the confidence of \( p_i, q_i \), we first need to establish a "reference standard" which will be used to assess the affinity of \( p_i \) and \( q_i \) to this standard. This affinity represents the confidence information we aim to obtain. Our local feature extractor can be implemented in various ways, such as using BRIEF (Binary Robust Independent Elementary Features)~\cite{calonder2010brief} descriptors or other deep learning-based descriptors~\cite{detone2018superpoint}. Here, we adopt a more efficient approach: for a given point \( x_i, y_i \), we define a small square region centered around it, and compute the normalized brightness distribution histogram of this region as its local pattern.

Due to the presence of both positive and negative terminals in lithium batteries, with slight differences in their local patterns, we use two colors in the figure: red for the positive terminal. To address this, we design a standard feature extractor that summarizes the local patterns of both the positive and negative terminals in an input X-ray image, calculating their average as the reference feature \( r_i \). Similarly, we use a patch feature extractor to extract the local features $u_i,v_i$ for \( p_i \) and \( q_i \) individually. Thus the corresponding confidence 
$\alpha_i, \beta_i$ can be achieved as follows:
\begin{equation}
\alpha_i = <\frac{v_i}{||v_i||}, \frac{r_i}{||r_i||}>,\quad \beta_i = <\frac{u_i}{||u_i||}, \frac{r_i}{||r_i||}>
\end{equation}

Once the confidence for each is captured as a measure of uncertainty, we naturally aggregate the pole estimates from both the \( p_i:\{x_i,y_i\} \) and \( q_i:\{x'_i,y'_i\} \).
\begin{equation}
\hat{x_i} = \frac{\alpha_i x_i + \beta_i x'_i}{\alpha_i + \beta_i},\quad
\hat{y_i} = \frac{\alpha_i y_i + \beta_i y'_i}{\alpha_i + \beta_i}
\end{equation}
where, $\hat{x_i} $ and $\hat{y_i} $ are the refined pole coordinates.

\section{Experimental Evaluation}
\subsection{Experimental setting}
We first apply OFAST to detect \( N \) corner points. Based on this, we determine the ROI and input it into HRNet. Finally, we use our confidence-based coordinate optimization module. We provide baseline results based on HRNet and compare the results with different choices of \( N \), corresponding to different numbers of corner points extracted by OFAST. 

\subsection{Evaluation Metric}

We adopt diverse metrics~\cite{yu2021heatmap} to compare different experimental settings including NME, PCK@0.5\%, PCK@1.0\% PCS@0.5\% and PCS@1.0\%.

Normalized Mean Error (NME) is used to evaluate the accuracy of predicted keypoints compared to the ground truth. It calculates the mean of the normalized errors across all samples and keypoints, with a lower NME indicating better performance.
\begin{equation}
NME = \frac{\sum_s \sum_i d_{err}^{s,i} / d_{ref}^i}{\sum_s \sum_i 1}
\end{equation}

where \( d_{err}^{s,i} \) is the error (distance) between the predicted and true coordinates for keypoint \( i \) in sample \( s \). \( d_{ref}^i \) is the reference distance for keypoint \( i \). Here we adopt the diagonal length of the ROI as reference distance.

PCK (Percentage of Correct Keypoints) calculates the proportion of keypoints that are within the acceptable error threshold (\( \theta \)), normalized by the total number of keypoints. It is often used to evaluate keypoint detection accuracy, with a higher PCK indicating better performance.

\begin{equation}
PCK_{\theta} = \frac{\sum_s \sum_i \delta (d_{err}^{s,i} / d_{ref}^s \leqslant \theta)}{\sum_s \sum_i 1}
\end{equation}


PCS (Prcentage of Correct Samples) calculates the proportion of samples where the maximum normalized error (among all keypoints) is less than or equal to the threshold \( \theta \). A higher \( PCS_{\theta}^{max} \) indicates better performance, as more samples have their maximum error within the acceptable range.
\begin{equation}
PCS_{\theta}^{max} = \frac{\sum_s \delta (\textbf{Max}_i [d_{err}^{s,i} / d_{ref}^s] \leqslant {\theta})}{\sum_s 1}
\end{equation}

\subsection{Numerical Result}
As shown in Tabe.~\ref{tab:performance_comparison}, the experimental results demonstrate that the addition of corner points significantly improves keypoint localization performance, especially when the number of corner points is increased. HRNet combined with corner points (particularly with $N=512$) consistently outperforms the baseline HRNet across all evaluation metrics. These improvements indicate that corner points provide valuable information that enhances the precision of keypoint detection, particularly for tasks requiring high accuracy.


\begin{table}[ht]
\caption{Performance comparison of different settings.}
\vspace{12pt}
\label{tab:performance_comparison}
\centering
\resizebox{\textwidth}{!}{
\begin{tabular}{lccccc}
\toprule
 \textbf{Method} & \textbf{NME(\%)}$\downarrow$ & \textbf{PCK@0.5\%}$\uparrow$ & \textbf{PCK@1.0\%}$\uparrow$ & \textbf{PCS@0.5\%}$\uparrow$ & \textbf{PCS@1.0\%}$\uparrow$ \\
\midrule
 HRNet (Baseline) & 0.658 & 0.421 & 0.818 & 0.239 & 0.965 \\
 HRNet + Corner points (N=128) & 0.624 & 0.473 & 0.836 & 0.303 & 0.978 \\
 HRNet + Corner points (N=256) & 0.611 & 0.486 & 0.847 & 0.337 & 0.978 \\
 HRNet + Corner points (N=512) & 0.582 & 0.519 & 0.865 & 0.388 & 0.991 \\
\midrule
 \large{$\Delta$ \textbf{Relative Gain}} & \large{\textbf{11.55\%}} & \large{\textbf{23.28\%}} & \large{\textbf{5.75\%}} & \large{\textbf{62.34\%}} & \large{\textbf{2.69\%}} \\
\bottomrule
\end{tabular}
}
\end{table}

\section{Conclusion}

In this paper, we propose a joint optimization model that integrates CNN-based heatmap regression with local gradient-based corner point detection. This approach significantly improves the localization of pole coordinates in X-ray images of lithium batteries. \textbf{\textit{Our findings suggest that, for keypoint extraction tasks using deep learning, combining traditional pixel gradient analysis methods with CNN-based heatmap regression is a promising strategy. This hybrid approach enhances both the accuracy and efficiency of the solution, leading to notable performance improvements.}}

\clearpage

{
\small
\bibliographystyle{IEEEtran}
\bibliography{reference}
}
\end{document}